\documentclass[sigconf]{acmart}

\AtBeginDocument{%
  \providecommand\BibTeX{{%
    \normalfont B\kern-0.5em{\scshape i\kern-0.25em b}\kern-0.8em\TeX}}}

\copyrightyear{2025}
\acmYear{2025}
\setcopyright{acmlicensed}
\acmConference[MM '25] {Proceedings of the 33rd ACM International Conference on Multimedia}{October 27--31, 2025}{Dublin, Ireland.}
\acmBooktitle{Proceedings of the 33rd ACM International Conference on Multimedia (MM '25), October 27--31, 2025, Dublin, Ireland}
\acmISBN{979-8-4007-2035-2/2025/10}
\acmDOI{10.1145/XXXXXX.XXXXXX}

\acmSubmissionID{1479}

\usepackage{booktabs}
\usepackage[skip=5pt]{caption}
\usepackage{algorithm}
\usepackage{multirow}
\usepackage{algorithmicx}
\usepackage{algpseudocode}
\usepackage{amsmath}
\usepackage{mathtools}
\usepackage{url}
\usepackage{balance}
\usepackage{makecell}

\usepackage{graphicx}

\usepackage{xcolor}
\usepackage{makecell}
\usepackage{colortbl}

\usepackage[capitalize]{cleveref}
\crefname{section}{Sec.}{Secs.}
\Crefname{section}{Section}{Sections}
\Crefname{table}{Table}{Tables}
\crefname{table}{Tab.}{Tabs.}

\usepackage{subcaption}

\usepackage{enumitem}

\makeatletter
\newcommand{\rmnum}[1]{\romannumeral #1}
\newcommand{\Rmnum}[1]{\expandafter\@slowromancap\romannumeral #1@}
\makeatother

\UseRawInputEncoding

\begin{document}

\title{Rethinking Occlusion in FER: A Semantic-Aware Perspective and Go Beyond}

\author{Huiyu Zhai}
\orcid{0009-0008-7498-8505}
\authornotemark[1]
\affiliation{%
  \institution{School of Computer Science and Engineering, University of Electronic Science and Technology of China}
  \city{Chengdu}
  \state{Sichuan}
  \country{China}
}
\email{202522081112@std.uestc.edu.cn}

\author{Xingxing Yang}
\authornote{These authors contributed equally to this work.}
\orcid{0009-0002-4522-5492}
\affiliation{%
  \institution{Department of Computer Science, Hong Kong Baptist University}
  \state{Hong Kong SAR}
  \country{China}
}
\email{csxxyang@comp.hkbu.edu.hk}

\author{Yalan Ye}
\orcid{0000-0001-5974-1717}
\authornote{Corresponding authors.}
\affiliation{%
    \institution{School of Computer Science and Engineering, University of Electronic Science and Technology of China}
  \city{Chengdu}
  \state{Sichuan}
  \country{China}
}
\email{yalanye@uestc.edu.cn}

\author{Chenyang Li}
\orcid{0000-0003-1212-4476}
\affiliation{%
    \institution{School of Computer Science and Engineering, University of Electronic Science and Technology of China}
  \city{Chengdu}
  \state{Sichuan}
  \country{China}
}
\email{202411081509@std.uestc.edu.cn}

\author{Bin Fan}
\orcid{0009-0004-7283-0983}
\affiliation{%
    \institution{School of Computer Science and Engineering, University of Electronic Science and Technology of China}
  \city{Chengdu}
  \state{Sichuan}
  \country{China}
}
\email{202421080407@std.uestc.edu.cn}

\author{Changze Li}
\orcid{0009-0003-1911-2061}
\affiliation{%
    \institution{School of Computer Science and Engineering, University of Electronic Science and Technology of China}
  \city{Chengdu}
  \state{Sichuan}
  \country{China}
}
\email{2022040908007@std.uestc.edu.cn}

\renewcommand{\shortauthors}{Huiyu Zhai, Xingxing Yang, Yalan Ye, Chenyang Li, Bin Fan, Changze Li}
\begin{abstract}
Facial expression recognition (FER) is a challenging task due to pervasive occlusion and dataset biases. 
Especially when facial information is partially occluded, existing FER models struggle to extract effective facial features, leading to inaccurate classifications. 
In response, we present ORSANet, which introduces the following three key contributions:
First, we introduce auxiliary multi-modal semantic guidance to disambiguate facial occlusion and learn high-level semantic knowledge, which is two-fold: 1) we introduce semantic segmentation maps as dense semantics prior to generate semantics-enhanced facial representations; 2) we introduce facial landmarks as sparse geometric prior to mitigate intrinsic noises in FER, such as identity and gender biases.
Second, to facilitate the effective incorporation of these two multi-modal priors, we customize a Multi-scale Cross-interaction Module (MCM) to adaptively fuse the landmark feature and semantics-enhanced representations within different scales.
Third, we design a Dynamic Adversarial Repulsion Enhancement Loss (DARELoss) that dynamically adjusts the margins of ambiguous classes, further enhancing the model's ability to distinguish similar expressions.
We further construct the first occlusion-oriented FER dataset to facilitate specialized robustness analysis on various real-world occlusion conditions, dubbed Occlu-FER.
Extensive experiments on both public benchmarks and Occlu-FER demonstrate that our proposed ORSANet achieves SOTA recognition performance.
Code is publicly available at \href{https://github.com/Wenyuzhy/ORSANet-master}{https://github.com/Wenyuzhy/ORSANet-master}.
\end{abstract}




\maketitle


\section{Introduction}\label{sec:introduction}

\begin{flushleft}
  ``\textit{Of Mountain Lu we cannot make out the true face, for we are lost in the heart of the very place.}'' \vspace{0.05in}
  \\\raggedleft{------ Su Shi, 1084} 
\end{flushleft}

\begin{flushleft}
  ``\textit{Fear not the floating clouds, but be at the highest level.}''
  \\\raggedleft{------ Wang Anshi, 1050} 
\end{flushleft}

The pursuit of high-level understanding in recognition tasks has driven significant advancements in the field of Facial Expression Recognition (FER)~\cite{li2020deep, wang2020suppressing, revina2021survey, zhu2023variance}, which aims to accurately recognize different facial expressions and has various applications, such as human-computer interaction (HCI)~\cite{abdat2011human, chowdary2023deep} and psychological research~\cite{ge2022facial, sajjad2023comprehensive, yu2024exploring}.
Most existing methods~\cite{zhao2021robust, zeng2022face2exp, chen2021cross} deal with this problem in an end-to-end manner. 
For example, Xue et al.~\cite{xue2021transfer} propose Transfer, which enhances facial expression analysis performance by leveraging local information perception and global information integration.
POSTER~\cite{zheng2023poster} leverages landmark features to guide the network's attention toward salient facial regions. 
Landmark features can explicitly model the geometric structure of facial expressions, making them less sensitive to noise factors such as skin color, gender, and background appearance. 
However, in real-world applications, FER often struggles with disturbances caused by various factors, such as identity, pose, illumination, scale-sensitivity, occlusion, and so on~\cite{kopalidis2024advances, liu2023learning, zhu2023variance}, where occlusion is one of the most challenging factors among them.

To address the occlusion issue, Zhao et al.~\cite{zhao2021robust} designed a local feature extractor and a channel-spatial modulator to enhance salient feature extraction.
Lee et al.~\cite{lee2023latent} proposed a mask-and-then-reconstruction framework by employing masks as occlusion information and then using a ViT-based~\cite{dosovitskiy2020image} reconstruction network to reconstruct the occluded regions.
However, these methods still treat occlusion and salient feature extraction brutally without auxiliary information guidance, which may not be effective and robust in various occlusion scenarios, especially in real-world applications.
A straightforward question arises: Are existing methods robust to occlusion conditions?

To investigate this question, our analysis reveals two core limitations in current methods:
\rmnum{1}, Semantic Understanding Failure. When the face is partially occluded by objects (\textit{e.g.}, glasses or hands), or extraneous faces appear in non-primary regions, conventional methods focus on less salient regions due to a lack of understanding of high-level semantic knowledge, leading to wrong classification, as illustrated in Fig.~\ref{fig:teaser}.
\rmnum{2}, Dataset Imbalance. The proportion of naturally occluded samples in mainstream datasets is relatively low, making it difficult for models to learn discriminative features under various occlusion conditions. Therefore, existing methods often use rectangular masks or random erasing to simulate occlusions; however, such artificial approaches differ significantly in feature distribution from real-world semantic occlusions (\textit{e.g.}, masks, hands), limiting their effectiveness in improving model generalization.

These limitations resonate with the reflections articulated in the poetry of Su Shi, a distinguished thinker from the Northern Song Dynasty of China.
In contrast, another poem by the reformer Wang Anshi suggests that understanding deepens when one adopts alternative perspectives. 
Motivated by this philosophical insight, auxiliary information and new objective functions should be introduced as explicit guidance, which 1) reduces the uncertainty of object detection and complements high-quality semantic knowledge details in occluded regions and 2) distinguishes small-scale target samples from the most easily confusable large-scale negative samples to improve the accuracy of classification decisions.

\begin{figure}[t]
    \begin{center}
       \includegraphics[width=0.9\linewidth]{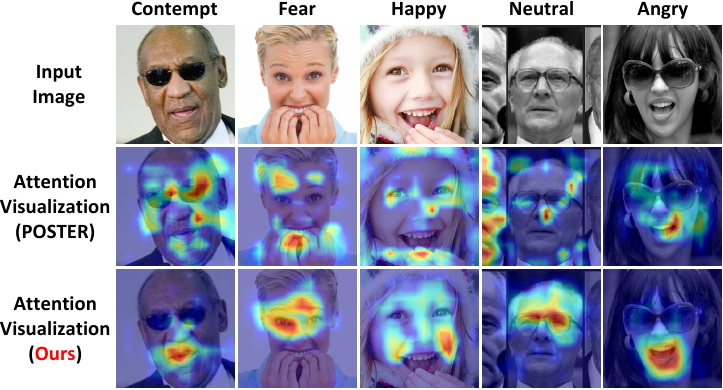} 
    \end{center}
    \caption{Samples in our Occlu-FER dataset. The last two rows show that POSTER~\cite{zheng2023poster} fails to extract salient facial features while our method accurately focuses on the key features of different expression categories.}
    \label{fig:present} 
    \vspace{-0.8em}
\end{figure}

To this end, we propose the Occlusion-Robust Semantic-Aware Network (ORSANet), which introduces semantic guidance to learn high-level semantic knowledge to deal with the challenging occlusion issue in FER.
Specifically, our semantic guidance is two-fold:
First, we introduce semantic segmentation maps as dense semantics prior extracted by a pre-trained facial segmentation model to align the facial feature space via the Spatial-Semantic Guidance Module (SSGM), learning high-level semantic information in occluded regions. Benefiting from the high-resolution semantic segmentation maps, more accurate local details can be captured for semantic-aware facial representation generation.
Second, considering the intrinsic noise in FER, such as identity and gender biases, which may interrupt models focusing on irrelevant features, leading to incorrect classification and unstable performance, we further introduce facial landmarks as sparse geometric prior to allocating generic facial component distributions, which filters out those noises and thus facilitating robust recognition.
To effectively incorporate both dense semantics prior and sparse geometric prior, we customize a Multi-scale Cross-interaction Module that incorporates a multi-scale interaction mechanism and a reintegration mechanism to adaptively fuse the landmark feature and semantics-enhanced facial features within different scales.
In addition, to deal with the dataset imbalance issue, we design a new objective function, dubbed Dynamic Adversarial Repulsion Enhancement Loss (DARELoss). It can enlarge the decision boundaries among highly similar expression categories by adaptively suppressing the most competitive negative class, significantly improving the discriminability of similar facial expressions.
Finally, we construct a new dataset to facilitate specialized robustness analysis on various occlusion conditions, dubbed Occlu-FER dataset. Some samples and performance comparisons in the Occlu FER dataset are provided in Fig.~\ref{fig:present}.
Our main contributions are as follows:

\begin{figure}
    \begin{center}
       \includegraphics[width=1.00\linewidth]{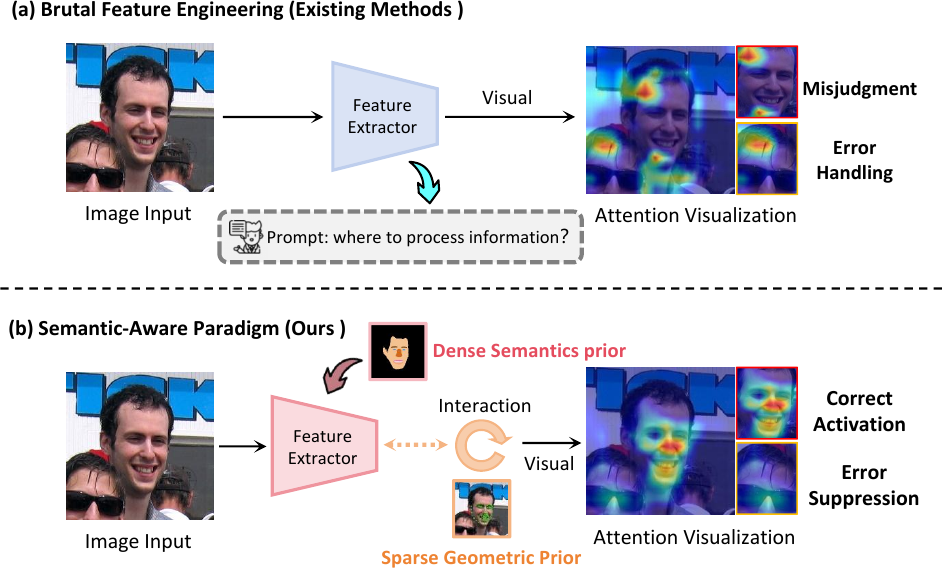} 
    \end{center}
    \caption{Motivation comparison. Existing methods (top) face dual challenges of error accumulation and misjudgment propagation under complex scenarios. In contrast, Our proposed ORSANet (bottom) introduces a semantic-aware mechanism, enabling a paradigm shift from ``passive error correction'' to ``proactive discrimination''.}
    \label{fig:teaser} 
    \vspace{-1.0em}
\end{figure}

\begin{itemize}[leftmargin=*]
    \item \textbf{New Method.} We propose ORSANet, which introduces both dense semantics prior (\textit{i.e.}, semantic segmentation maps) and sparse geometric prior (\textit{i.e.}, facial landmarks) as explicit semantic guidance to learn high-level semantic knowledge to deal with the challenging occlusion issue, achieving SOTA performance on several widely used benchmarks.
    \item \textbf{New Interaction Mechanism.} We customize a Multi-scale Cross-interaction Module (MCM) to effectively fuse the semantics-enhanced facial representations with landmark features, which disentangles expression-related features from irrelevant attributes through multi-scale interaction and reintegration manners, thereby mitigating the impact of intrinsic noise factors in FER.
    \item \textbf{New Loss Function.} We propose a Dynamic Adversarial Repulsion Enhancement Loss (DARELoss) for facilitating the learning of the complex samples, which expands the decision boundaries among highly similar categories, significantly improving the discriminability of facial expressions. Furthermore, this loss function shows great potential in general classification tasks.
    \item \textbf{New Benchmark.} We construct the Occlu-FER dataset, the first dataset tailored for FER in various real-world occlusion conditions, which could serve as a new benchmark for this challenging task.
\end{itemize}

\section{Related Work}\label{sec:related work}
\subsection{Deep Learning in FER}
Since deep learning dominates in visual affective analysis~\cite{wu2024bridging, zhao2024hawkeye}, convolutional neural networks (CNNs) ~\cite{gu2018recent, li2024temporal} have been widely applied to facial expression recognition tasks, significantly improving the performance of FER~\cite{vo2020pyramid, tao2023freq}.
Sang et al.~\cite{sang2018discriminative} focused on reducing intra-class variations in deep facial expression features and introduced a densely connected convolutional network~\cite{huang2017densely} for FER, while Savchenko et al~\cite{savchenko2021facial} explored the application of lightweight CNNs in FER tasks. 
However, as image features are inherently sensitive to factors such as skin tone, gender, and background appearance, relying solely on single-image information is insufficient to comprehensively tackle the challenges of FER tasks~\cite{yi2014facial, sajjad2018facial, kumar2023automatic, pan2023progressive, jing2023styleedl}.
With the advent of highly accurate facial landmark detectors~\cite{wang2020deep, wang2021face, jin2021pixel}, researchers have increasingly focused on utilizing landmark information to enhance FER performance. 
POSTER~\cite{zheng2023poster}, a representative model in this category, employed a synergistic guidance mechanism that integrated landmarks and image features.
Nevertheless, they ignored the potential risks of facial occlusion information and failed to fully exploit the advantages of landmarks' sparsity. 
As a result, they illustrate limited generalization capability in real-world scenarios.

\subsection{Uncertainly in Real-World FER}
Facial data in the real-world environment is uncontrollable, and it is necessary not only to address expression variations but also to manage issues such as occlusion and redundant faces, which causes instability in feature extraction. 
Pan et al.~\cite{pan2019occluded} trained two networks using occluded and non-occluded facial images and guided the learning process of the occluded network using the non-occluded network. 
At the same time, Wang et al.'s RAN~\cite{wang2020region} model divided the facial region into multiple small patches and adaptively captured the effects of occlusion and pose variations on FER. 
However, such methods often involve complex model structures and cumbersome training processes.
In addition, most existing FER studies~\cite{ruan2021feature, mao2024poster++} often overlooked the semantic interplay between occluded regions and facial context, resulting in their inability to effectively handle different types of occlusions. 
Therefore, more robust and efficient approaches are urgently needed to enhance FER generalization in real-world scenarios.

\subsection{Facial Parsing Development}
Facial parsing is a semantic segmentation task~\cite{long2015fully} that aims at assigning pixel-level labels to facial images to accurately distinguish key facial regions\cite{guo2018residual, zhou2015interlinked, lin2021roi}. 
AGRNet~\cite{te2021agrnet} and EAGRNet~\cite{te2020edge} employed graph-based representations to establish relationships among different facial components and leveraged edge information for parsing. 
DML-CSR~\cite{zheng2022decoupled} explored multi-task learning to address the challenge of noisy labels. 
Meanwhile, SegFace~\cite{narayan2024segface} introduced a lightweight Transformer decoder that integrated learnable class-specific tokens to achieve independent class modeling. 
Accurate semantic recognition of different facial regions is crucial for various applications, particularly in FER tasks, where it provides more stable and precise feature support.

\section{Method}\label{sec:method}

\subsection{Preliminaries: Incorporating of Dense Semantics Prior and Sparse Geometric Prior}
As mentioned earlier, when the face is partially occluded by objects that closely resemble facial features, or extraneous faces appear in non-primary regions, as illustrated in Fig.~\ref{fig:present}, brutally extracting features from both clear and occluded regions are suboptimal due to ambiguity between the target object and background, content and occlusion, leading to inaccurate classification.
A common approach to dealing with such a challenge is incorporating a latent variable $Z$. The posterior probability for the prediction $Y$ could be modeled as a conditional Variational Auto-encoder \cite{kingma2013auto, yang2023cooperative} given the input occluded facial expression image $X_{\text{N}}$ as:
\begin{small}
\begin{equation}
p\left(Y | X_{\text{N}}\right)=\int p\left(Y | Z, X_{\text{N}}\right) \cdot p\left(Z | X_{\text{N}}\right) dZ.
\end{equation}
\end{small}

There are multiple choices for the latent variable $Z$, which functions to learn high-level semantic knowledge to disambiguate the recognition of occluded regions. Intuitively, a semantic segmentation map could be introduced and employed as such a latent variable $Z$ to bring in additional semantic information for the classification task, which serves as a dense semantics prior.
Meanwhile, considering intrinsic noises in FER, such as identity and gender, which are content-related, we also utilize facial landmarks as a sparse geometric prior. This will clearly guide the network to focus more on expression-related features and filter out content-related noises.
In specific, we predict semantic segmentation maps (\textit{i.e.}, $X_{\text{N2S}}$) and facial landmarks (\textit{i.e.}, $X_{\text{N2L}}$) via a pre-trained segmentation model~\cite{narayan2024segface} and a facial landmark detection model~\cite{chen2021pytorch}, respectively. Both of them serve as the latent variable $Z$:
\begin{small}
\begin{align}
    \label{eq:domain translation description}
    p\left(Y | X_{\text{N}}\right) &=\int p\left(Y | X_{\text{N2S}}, X_{\text{N2L}}, X_{\text{N}}\right) \cdot p\left(X_{\text{N2S}}, X_{\text{N2L}} | X_{\text{N}}\right) dX_{\text{N2G}}.
\end{align}
\end{small}
Our consideration is two-fold:

(i) Multi-modal auxiliary information is introduced. Compared with a unimodal representation (\textit{e.g.}, latent spectrum translation used in \cite{yang2023cooperative}), we introduce both high-resolution semantic segmentation maps and sparse geometric landmarks, making it possible to capture more accurate local object details by the dense semantics prior and meanwhile, reduce content-related noises by the sparse geometric prior. Thus, the ambiguity between the target object and background, content and occlusion, can be addressed.

\begin{figure*}[ht]
    \vspace{-1.0mm}
    \begin{center}
        \includegraphics[width=0.95\textwidth]{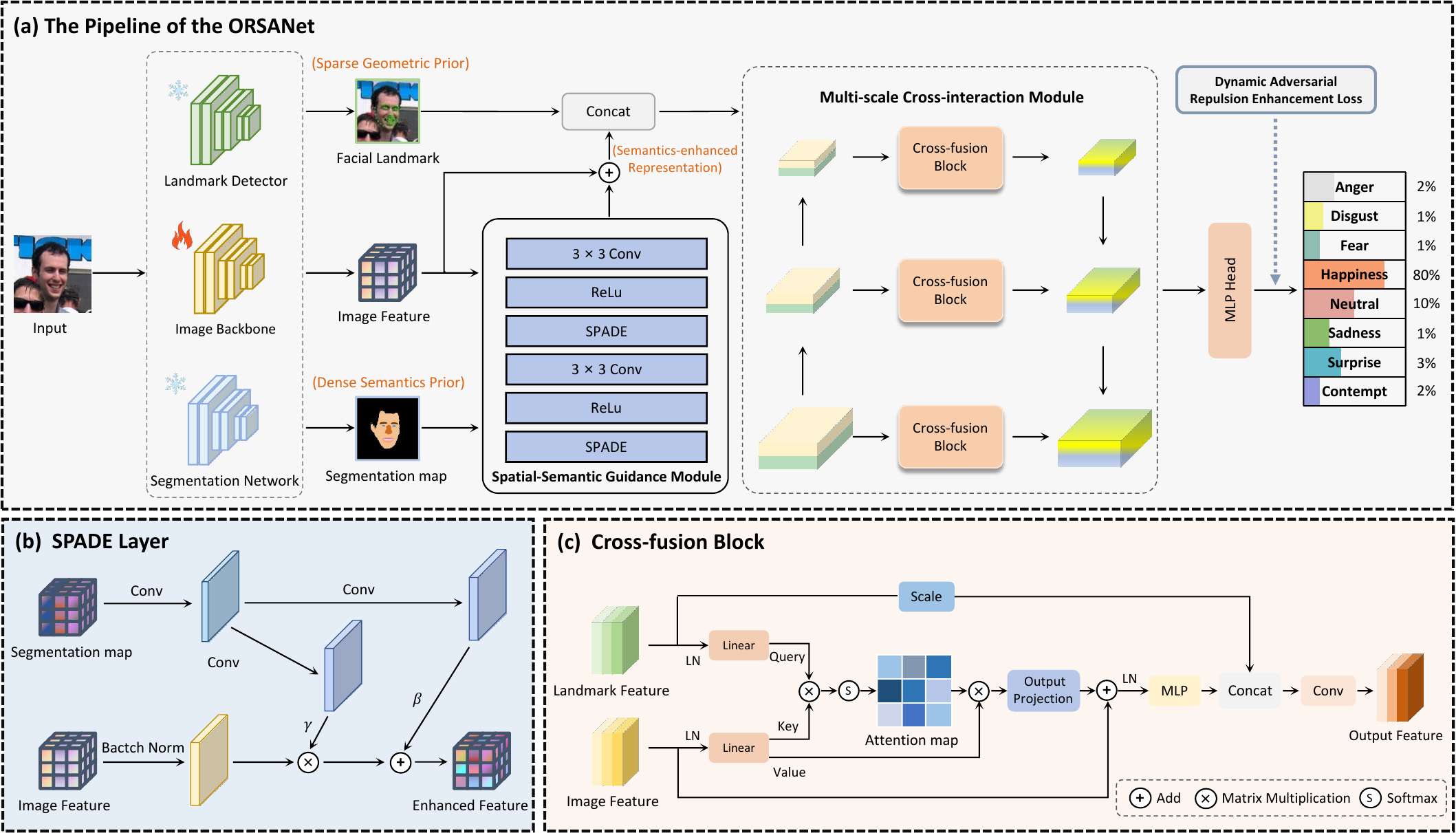}
    \end{center}
    \caption{Illustration of ORSANet. 
    (a) illustrates the overall pipeline. The sparse geometric prior and dense semantics prior are generated from a pre-trained landmark detector~\cite{chen2021pytorch} and a pre-trained semantic segmentation network~\cite{narayan2024segface}. Image features extracted from the trainable image backbone first interact with the dense semantics prior through the spatial-semantic guidance module (SSGM). Then, semantics-enhanced representations are concatenated with the sparse geometric prior to be fed into the multi-scale cross-interaction module (MCM) to disentangle expression-related features from irrelevant attributes. (b) shows the details of the SPADE~\cite{park2019semantic} in SSGM. (c) shows the details of the cross-fusion block in MCM.}
    \label{fig:pipeline}
    \vspace{-5pt}
\end{figure*}

(ii) Information Fusion benefits all tasks.
With both two priors introduced from external pre-trained models, multi-modal knowledge learned from each task domain can be complementary to others.
Thus, multi-modal information fusion is crucial. To this end, we customize a multi-scale cross-interaction module (MCM) that encourages efficient information integration between these two priors, which resolves ambiguities for each task.

\subsection{Method Overview}
As shown in Fig.~\ref{fig:pipeline}, ORSANet consists of three main components: feature extraction backbone, spatial-semantic guidance module (SSGM) for aligning facial spatial features, and multi-scale cross-interaction module (MCM) for feature fusion.
Firstly, the input facial images ${X}_\text{in}$ are fed into the backbone network to generate image features, facial landmarks, and segmentation maps.
\begin{small}
\begin{equation}
\label{eq:backbone}
    \begin{aligned}
    \mathbf{X}_{img}, \mathbf{X}_{seg}, \mathbf{X}_{lm} = G_{img}(\mathbf{X}_{in};\theta), G_{seg}(\mathbf{X}_{in}), G_{lm}(\mathbf{X}_{in}).
    \end{aligned}
\end{equation}
\end{small}
Then, we utilize segmentation maps as dense semantics prior to enhance the image feature as semantics-enhanced representations ${\hat{\mathbf{X}}}_{img}$ via SSGM.
Subsequently, ${\hat{\mathbf{X}}}_{img}$ are concatenated with the extracted landmark map $\mathbf{X}_{lm}$ (\textit{i.e.}, sparse geometric prior) and fed into MCM.
To effectively fuse these two multi-modal features, we design a multi-scale fusion mechanism and a reintegration mechanism in MCM, which adaptively fuse semantics-enhanced facial representations with landmark information to mitigate intrinsic noises.
Finally, the features ${\mathbf{X}}_{out}$ generated by MCM are fed into a classifier to predict the final facial expression category. 
\begin{small}
\begin{equation}
\label{eq:fusion}
    {\hat{\mathbf{X}}}_{img} = SSGM(\mathbf{X}_{img}, \mathbf{X}_{seg}), 
    {\mathbf{X}}_{out} = MCM({\hat{\mathbf{X}}}_{img}, \mathbf{X}_{lm}).
\end{equation}
\end{small}

\subsection{Spatial-Semantic Guidance Module}
The spatial-semantic guidance module (SSGM) integrates facial image features and semantic segmentation maps through a dual-stage Spatially-Adaptive Normalization (SPADE)~\cite{park2019semantic} enhancement unit for semantics-enhanced facial representations generation. 
The purpose of the first stage is to achieve coarse-grained spatial alignment. The facial features $\mathbf{X}_{img}$ and segmentation information $\mathbf{X}_{seg}$ are fed into the initial SPADE layer, where the pixel-wise scaling factor \textbf{$\gamma$} and the offset factor \textbf{$\beta$} generated from segmentation maps are used for spatially adaptive modulation, as described in Eq.~\ref{eq:spade}, aligning facial features with the overall semantic layout. 
\begin{small}
\begin{equation}
\label{eq:spade}
    \begin{aligned}
    & \mathbf{X}_{seg}^{\prime} = ReLU(Conv(\mathbf{X}_{seg})), \\
    & \mathbf{X}_{img}^{\prime} = Norm(\mathbf{X}_{img})\cdot Conv_\gamma(\mathbf{X}_{seg}^{\prime}) + Conv_\beta(\mathbf{X}_{seg}^{\prime}).
\end{aligned}
\end{equation}
\end{small}

The second stage focuses on fine-grained semantics information fusion, where the SPADE layer is employed again to further refine the feature representation. 
This step extracts higher-order semantics information from facial segmentation features at a deeper level, recovering discriminative cues in occluded regions while emphasizing crucial facial information. 
The module follows a progressive ``coarse-to-fine'' processing paradigm~\cite{xu2023dynamic, lu2024coarse}, to achieve a deep enhancement of external dense semantics prior knowledge and internal image feature, ultimately outputting semantic-enhanced facial feature representation.

\subsection{Multi-scale Cross-interaction Module}
The multi-scale cross-interaction module (MCM) adopts a hierarchical progressive feature fusion architecture~\cite{lin2017feature, yang2024hyperspectral}, consisting of stacked multi-level cross-fusion blocks (CFBs). 
The fusion module incrementally integrates multi-scale feature information through an adaptive feature interaction strategy at each level. 
Fig.~\ref{fig:pipeline}(c) illustrates the details of the cross-fusion block (CFB). A Cross-attention mechanism~\cite{chen2021cross} is firstly introduced to promote feature interaction between landmarks and image representations, while a reintegration mechanism is customized to fully exploit the sparse geometric potential of landmark features.
To ensure the network focuses on expression-relevant facial regions, we map the input landmark features $\mathbf{X}_{lm}\in\mathbb{R}^{P \times D}$ into a query matrix $\mathbf{Q}_{lm}$ via a linear transformation, while the facial image features ${\hat{\mathbf{X}}}_{img}\in\mathbb{R}^{P \times D}$ are mapped into key matrix $\mathbf{K}_{img}$ and value matrix $\mathbf{V}_{img}$: 
\begin{small}
\begin{equation}
\label{eq:matrix}
    \begin{aligned}
    \mathbf{Q}_{lm} = \mathbf{W}_q \cdot \mathbf{X}_{lm}, \mathbf{K}_{img} = \mathbf{W}_k \cdot {\hat{\mathbf{X}}}_{img}, \mathbf{V}_{img} = \mathbf{W}_v \cdot {\hat{\mathbf{X}}}_{img},
    \end{aligned}
\end{equation}
\end{small}
where $\mathbf{W}_q$, $\mathbf{W}_k$ and $\mathbf{W}_v$ $\in\mathbb{R}^{D \times D}$. We calculate the similarity between the $\mathbf{Q}_{lm}$ and the $\mathbf{K}_{img}$ via matrix multiplication, followed by Softmax normalization to generate attention maps. 
Then multiply with the $\mathbf{V}_{img}$ to obtain attention features:
\begin{small}
\begin{equation}
\label{eq:softmax}
    \begin{aligned}
    \mathbf{F}_{att} = Softmax\left(\frac{\mathbf{Q}_{lm} \cdot \mathbf{K}_{img}^T}{\sqrt{\mathbf{d}_i}}\right) \mathbf{V}_{img},
    \end{aligned}
\end{equation}
\end{small}
where $\mathbf{d}$ is a parameter that adaptively scales the matrix multiplication. The weighted fused features are further processed through an MLP to enhance feature representation capability:
\begin{small}
\begin{equation}
\label{eq:mlp}
    \begin{aligned}
    \mathbf{X}_{fuse} = MLP(Norm({\hat{\mathbf{X}}}_{img} + \mathbf{F}_{att})).
    \end{aligned}
\end{equation}
\end{small}
Finally, to further reduce facial interference caused by identity, gender, and age variations, an adaptive learning factor $s$ is introduced to further integrate landmarks and facial features, and finally output them through a $1 \times 1$ convolution:
\begin{small}
\begin{equation}
\label{eq:concat}
    \begin{aligned}
    \mathbf{X}_{out} = Conv(Concat(\mathbf{X}_{fuse} + s \cdot \mathbf{X}_{lm})).
    \end{aligned}
\end{equation}
\end{small}

The proposed MCM leverages sparse geometric prior from landmark by a multi-scale interaction mechanism and a reintegration mechanism, not only effectively mitigating the impact of intrinsic noises to disentangle expression-related features from irrelevant attributes in FER tasks, but also accommodating the scale sensitivity requirements of fine-grained FER tasks.

\subsection{DARELoss}
Motivated by the phenomenon that there often exist ``confusing'' negative classes that are highly similar to the target class~\cite{zeng2022face2exp, li2020deep}, where some categories may have response values close to the target category in the logits space, leading to ambiguous classification decision boundaries, we propose a new objective function that aims to distinguish the target class from the most easily confusable negative samples, thereby improving the accuracy of classification decisions, dubbed dynamic adversarial repulsion enhancement loss (DARELoss):

\begin{small}
\begin{equation}
\label{eq:DARELoss}
    \begin{aligned}
    \mathcal{L}_{dare}=-\log\frac{\mathbf{e}^{\mathbf{z}_x}}{\mathbf{e}^{\mathbf{z}_x}+\mathbf{e}^{\mathbf{z}_{y}^{\prime}}},
    \end{aligned}
\end{equation}
\begin{equation}
\label{eq:dynamic}
    \alpha = 1 - P(x),
    \mathbf{z}_{y}^{\prime} = \alpha \cdot \mathbf{z}_{y}+ \mathbf{z}_{y},
\end{equation}
\end{small}
where $\mathbf{z}_x$ represents the logits of the target class, and $\mathbf{z}_y$ denotes the logits of the maximum response negative class excluding the target class. 
By suppressing the competitive negative class, we compel the model to learn more discriminative feature representations. 
Furthermore, as shown in Eq.~\ref{eq:dynamic}, we design a dynamic confidence-aware mechanism that uses the predicted probability of the target class to assess the confidence of the model. 
When the model has low confidence in the target category, it enhances the contrast with the target class by increasing the logits of the competitive classes; conversely, when the model has high confidence, the penalty is reduced to avoid an overfitting problem. 
Finally, integrated with the commonly used cross-entropy loss, the final loss function is defined as follows: 
\begin{small}
\begin{equation}
    \label{eq:loss}
    \begin{aligned}
    \mathcal{L}=\lambda_1\mathcal{L}_{ce}+\lambda_2\mathcal{L}_{dare},
    \end{aligned}
\end{equation}
\end{small}
where $\lambda_{1}$ and $\lambda_{2}$ are trade-off weights tuned to balance the contributions of each loss term.

\section{Experiments}
\subsection{Datasets}
We conduct a comprehensive evaluation and comparison using RAF-DB~\cite{li2017reliable}, AffectNet~\cite{mollahosseini2017affectnet} and our constructed Occlu-FER dataset.

\textbf{RAF-DB.} The RAF-DB~\cite{li2017reliable} is a large-scale facial expression recognition dataset comprising 29,672 real-world facial images, covering seven basic emotion categories (neutral, happy, sad, surprised, fearful, disgusted, and angry). 
Most of the samples contain at least one type of interference factor, exhibiting variations such as occlusion, multi-pose and diverse resolutions.

\textbf{AffectNet.} AffectNet~\cite{mollahosseini2017affectnet} contains more than 400,000 facial images related to emotional words crawled from the internet, and is currently the largest public facial expression dataset.
In addition to the seven basic emotion labels, this dataset includes "contempt" as an additional emotion category, and provides continuous dimension annotations for valence and arousal.

\begin{table}[h]
    \vspace{-0.5mm}
    \caption{The detailed presentation of the Occlu-FER dataset, which includes eight basic emotion categories.}
    \label{tab:Occlu-FER}
    \centering
    \renewcommand{\arraystretch}{1.1} 
    \resizebox{1.00\linewidth}{!}{
        \begin{tabular}{l|cccccccc|c}
            \toprule
            \textbf{Label} & \textbf{AN} & \textbf{DI} & \textbf{FE} & \textbf{HA} & \textbf{NE} & \textbf{SA} & \textbf{SU} & \textbf{CO} &\textbf{Sum}\\ 
            \midrule
            Train & 669 & 533 & 927 & 1130 & 1114 & 960 & 1125 & 380 & 6838 \\
            Valid & 86  & 69  & 136 & 133  & 144  & 127 & 148  & 37  & 880 \\
            \bottomrule
        \end{tabular}
    }
    \vspace{-0.8em}
\end{table}

\begin{table*}[t]
    \centering
    \caption{Comparison results with SOTA FER methods on RAF-DB~\cite{li2017reliable} (8cls) and AffectNet~\cite{mollahosseini2017affectnet} (7cls). All metrics represent the overall prediction accuracy (\%) of the entire validation dataset. The best results are highlighted in bold.}
    \label{tab:results}
    \renewcommand{\arraystretch}{1.0}
    \resizebox{0.75\textwidth}{!}{ 
        \begin{tabular}{l|l|l|cccc}
        \toprule
        \textbf{Category} & \textbf{Methods} & \textbf{Venue} & \textbf{RAF-DB} & \textbf{AffectNet (7cls)} & \textbf{AffectNet (8cls)} \\
        \midrule
        
        \multirow{8}{*}{Natural Methods} 
        & FDRL~\cite{ruan2021feature}      & CVPR 2021   & 89.47   & --     & --     \\
        & TransFER~\cite{xue2021transfer}  & ICCV 2021   & 90.91   & 66.23  & --     \\
        & Face2Exp~\cite{zeng2022face2exp} & CVPR 2022   & 88.54   & 64.23  & --     \\
        & EAC~\cite{zhang2022learn}        & ECCV 2022   & 89.99   & 65.32  & --     \\
        & POSTER~\cite{zheng2023poster}    & ICCV 2023   & 92.05   & 66.17  & 62.05  \\
        & MMATrans~\cite{liu2024mmatrans}  & TII 2024    & 89.67   & 64.89  & --     \\
        & POSTER V2~\cite{mao2024poster++} & PR 2024     & \cellcolor{gray!20}\underline{92.21} & 66.20  & \cellcolor{gray!20}\underline{62.37} \\
        & COA~\cite{cao2025co}             & TCSVT 2025  & 91.13   & 66.00  & 62.19  \\
        
        \midrule
        
        \multirow{3}{*}{Multi-modal}
        & FER-former~\cite{li2024fer}      & TMM 2024    & 91.30   & --     & --     \\
        & CLEF~\cite{yang2024robust}       & CVPR 2024   & 91.46   & 65.76  & 62.13  \\
        & CLIPER~\cite{li2024cliper}       & ICME 2024   & 91.61   & \cellcolor{gray!20}\underline{66.29} & 61.98  \\
        
        \midrule
        
        \multirow{5}{*}{Occlusion}
        & RAN~\cite{wang2020region}              & TIP 2020    & 88.90   & --      & --   \\
        & EfficientFace~\cite{zhao2021robust}    & AAAI 2021   & 88.36   & 63.70   & 59.89 \\
        & MAPNet~\cite{ju2022mask}               & ICASSP 2022      & 87.26   & 64.09   & --    \\
        & Latent-OFER~\cite{lee2023latent}       & ICCV 2023   & 89.60   & 63.90   & --     \\
        & \textcolor{red}{\textbf{ORSANet}} \textit{(ours)}    & --  & \cellcolor{blue!25}\textbf{92.28} & \cellcolor{blue!25}\textbf{66.69} & \cellcolor{blue!25}\textbf{62.95}  \\ 
        
        \bottomrule
        
        \end{tabular}
    }
    \vspace{-5pt}
\end{table*}

\textbf{Occlu-FER.} To facilitate specialized robustness analysis on various real-world occlusion conditions, we construct the first occlusion dataset named Occlu-FER, which focuses on partial facial occlusion and extraneous face interference in real-world scenarios.
This dataset covers eight basic emotion categories, as described in Tab.~\ref{tab:Occlu-FER}. 
The training dataset consists of 6838 images, and the validation dataset includes 880 images. 
The image sources of the dataset include both occlusion samples from public in-the-wild datasets and real-world facial photographs collected from the internet.

The Occlu-FER dataset not only provides substantial experimental support for the training and validation of our model but also serves as a crucial benchmark for future research in related fields.

\subsection{Implementation Details}
In the feature extraction backbone network, we employ the pre-trained SegFace~\cite{narayan2024segface} as the semantic segmentation generator and select the pre-trained MobileFaceNet~\cite{chen2021pytorch} as the facial landmark detector. The weights of both networks are frozen during the training process to ensure the accurate extraction of relevant features. Finally, for the image backbone, we use IR50~\cite{deng2019arcface, guo2016ms} to extract facial features.
We set the learning rate to $1e^{-4}$ and adopt a batch size of 20, training the model for 400 epochs using the Adam optimizer~\cite{kingma2014adam}. For the loss function, we set parameter $\lambda_{1}$ to 1 and $\lambda_{2}$ to 0.1. The experimental source code is implemented with Pytorch, and the models are trained with a single NVIDIA RTX 3090.

\begin{table}[t]
    \small
    \centering
    \caption{Evaluation results of artificially simulated occlusion scenarios (10-30\%) on the RAF-DB~\cite{li2017reliable} validation dataset.}
    \label{tab:artificial}
    \setlength{\tabcolsep}{2mm}
    \renewcommand{\arraystretch}{1.3}
    \resizebox{1.0\linewidth}{!}{ 
        \begin{tabular}{l|l|cccc}
        \toprule
        \textbf{Category} & \textbf{Methods} & \textbf{Original} & \textbf{10\%} & \textbf{20\%} & \textbf{30\%} \\
        \midrule
        
        \multirow{5}{*}{Natural} 
        & FDRL~\cite{ruan2021feature}          & 89.47   & 88.36   & 85.36   & 81.22   \\
        & Face2Exp~\cite{zeng2022face2exp}     & 88.54   & 88.12   & 84.31   & 80.77   \\
        & EAC~\cite{zhang2022learn}            & 89.99   & 89.05   & 85.57   & 81.32   \\
        & POSTER~\cite{zheng2023poster}        & 92.05   & \cellcolor{gray!20}\underline{89.76} & 86.86   & \cellcolor{gray!20}\underline{82.53}   \\
        & POSTER V2~\cite{mao2024poster++}     & \cellcolor{gray!20}\underline{92.21} & 89.52   & \cellcolor{gray!20}\underline{86.95}   & 82.16   \\
        \midrule
        
        \multirow{2}{*}{Multi-modal}
        & CLEF~\cite{yang2024robust}           & 91.46   & 89.14   & 85.78   & 81.58   \\
        & CLIPER~\cite{li2024cliper}           & 91.61   & 88.56   & 85.94   & 81.74   \\
        \midrule
        
        \multirow{2}{*}{Occlusion}
        & EfficientFace~\cite{zhao2021robust}  & 88.36   & 87.93   & 84.79   & 80.93   \\
        & \textcolor{red}{\textbf{ORSANet}} \textit{(ours)} & \cellcolor{blue!25}\textbf{92.28} & \cellcolor{blue!25}\textbf{90.51} & \cellcolor{blue!25}\textbf{87.87} & \cellcolor{blue!25}\textbf{84.02}   \\ 
        
        \bottomrule
        \end{tabular}
    }
    \vspace{-1.8em}
\end{table}

\subsection{Comparison With State-of-the-Art}

\textbf{Results on Public In-the-Wild Datasets.}
We compare with SOTA methods in recent years under the same dataset settings, including FDRL~\cite{ruan2021feature}, TransFER~\cite{xue2021transfer}, Face2Exp~\cite{zeng2022face2exp}, EAC~\cite{zhang2022learn}, POSTER~\cite{zheng2023poster}, MMATrans~\cite{liu2024mmatrans}, POSTER V2~\cite{mao2024poster++}, COA~\cite{cao2025co}, FER-former~\cite{li2024fer}, CLEF~\cite{yang2024robust}, CLIPER~\cite{li2024cliper}, RAN~\cite{wang2020region}, EfficientFace~\cite{zhao2021robust}, MAPNet~\cite{ju2022mask}, Latent-OFER~\cite{lee2023latent}. 
We provide the quantitative results in Tab.~\ref{tab:results}, which indicates that our ORSANet not only optimizes for occlusion scenes but also achieves the best performance in various datasets: 
On the RAF-DB dataset, ORSANet achieves a recognition accuracy of 92.28\%, surpassing the previous state-of-the-art method POSTER V2~\cite{mao2024poster++} and significantly outperforming CLIPER based on vision-language alignment learning (+0.67\%), which strongly verifies the effectiveness of our chosen auxiliary semantic priors and guidance strategies.
Meanwhile, ORSANet also demonstrates outstanding performance on the more challenging AffectNet dataset. 
Among the 7-class basic expression recognition tasks, it reaches an accuracy of 66.69\%, outperforming Latent-OFER~\cite{lee2023latent} (63.90\%)-a model also designed for occlusion scenarios-by 2.79 percentage points. Furthermore, ORSANet maintains a leading position in the 8-class extended tasks, achieving an accuracy of 62.95\%, further proving the robustness and advancement of our approach in complex scenes.
More results are provided in the Supplementary Materials.

\begin{table}[t]
    \small
    \centering
    \caption{Experimental results in real-world occlusion scenarios, including RAF-DB (occlu) and Occlu-FER.}
    \label{tab:real-occlu}
    \renewcommand{\arraystretch}{1.2}
    \resizebox{0.98\linewidth}{!}{
        \begin{tabular}{l|l|cc}
        \toprule
        \textbf{Category} & \textbf{Methods} & \textbf{RAF-DB (occlu)} & \textbf{Occlu-FER} \\
        \midrule
        
        \multirow{5}{*}{Natural} 
        & FDRL~\cite{ruan2021feature}          & 85.03   & 65.86  \\
        & Face2Exp~\cite{zeng2022face2exp}     & 84.53   & 65.79  \\
        & EAC~\cite{zhang2022learn}            & 85.46   & 66.36  \\
        & POSTER~\cite{zheng2023poster}        & 86.82   & 66.82  \\
        & POSTER V2~\cite{mao2024poster++}     & \cellcolor{gray!20}\underline{87.02}   & 66.09  \\
        \midrule
        
        \multirow{2}{*}{Multi-modal}
        & CLEF~\cite{yang2024robust}           & 86.16   & 66.74   \\
        & CLIPER~\cite{li2024cliper}           & 86.68   & \cellcolor{gray!20}\underline{67.61}  \\
        \midrule
        
        \multirow{2}{*}{Occlusion}
        & EfficientFace~\cite{zhao2021robust}  & 85.39   & 65.96  \\
        & \textcolor{red}{\textbf{ORSANet}} \textit{(ours)}     & \cellcolor{blue!25}\textbf{87.75}   & \cellcolor{blue!25}\textbf{68.07}  \\ 
        
        \bottomrule
        \end{tabular}
    }
    \vspace{-1.0em}
\end{table}

\textbf{Results on Artificial Occlusion.}
To evaluate the recognition performance of the model under occlusion scenarios, we conduct controlled experiments by applying artificial masks to simulate three occlusion levels-10\%, 20\%, and 30\%-on the RAD-DB~\cite{li2017reliable} validation dataset. 
The quantitative results in Tab~\ref{tab:artificial} indicate that ORSANet consistently achieves the highest accuracy under the occlusion, ranging from 10\% (90.51\%) to 30\% (84.02\%). 
Moreover, as the level of occlusion increases, ORSANet demonstrates the smallest performance drop, and its superiority becomes increasingly evident compared to other methods under equivalent occlusion conditions.

\begin{figure}[t]
    \begin{center}
       \includegraphics[width=0.98\linewidth]{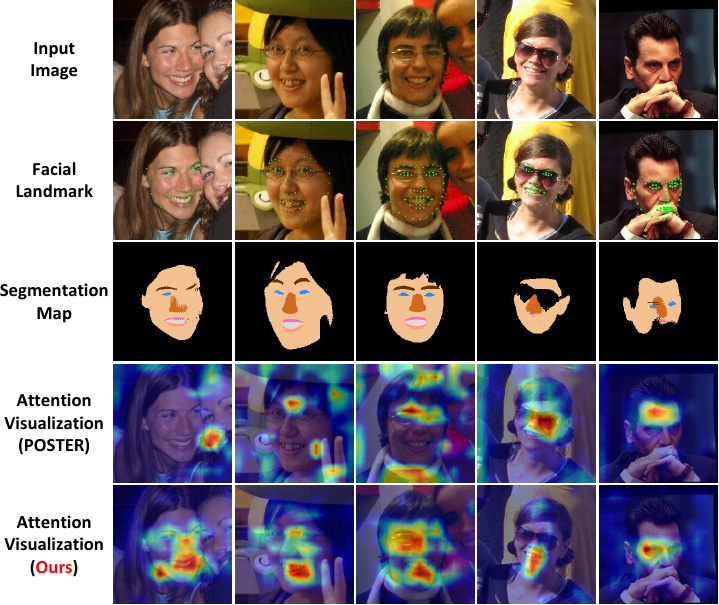} 
    \end{center}
    \caption{Visualization of facial expression. Including input image, facial landmark, segmentation map, and attention visualization (Ours ORSANet and POSTER~\cite{zheng2023poster}).}
    \label{fig:visual} 
    \vspace{-2.0em}
\end{figure}

\textbf{Results on Real-world Occlusion.}
To evaluate the performance of the model in real-world occlusion scenarios, we compare the performance of ORSANet with several SOTA methods on RAF-DB (occlu) and Occlu-FER using the same experimental setup. 
RAF-DB (occlu) comprises occluded samples from the RAF-DB~\cite{li2017reliable} validation dataset. 
As shown in Tab.~\ref{tab:real-occlu}, our model achieves the best results on both datasets.
For Occlu-FER, ORSANet reaches the highest accuracy of 68.05\%, significantly outperforming all compared methods except CLIPER~\cite{yang2024robust}.
A reasonable explanation for CLIPER's ~\cite{yang2024robust} comparable performance is that, unlike other datasets, Occlu-FER contains more contextual semantic cues-such as scene elements and body gestures-allowing CLIPER to leverage its multi-modal design to capture richer text description information and supervise the recognition process.
On RAF-DB (occlu), we directly evaluate the occluded samples using the model weights trained on the original dataset. ORSANet achieves the highest accuracy of 87.75\%, demonstrating excellent generalization performance.

\textbf{Qualitative Evaluation.}
To intuitively verify the decision-making mechanism of the model under occluded scenarios, Fig.~\ref{fig:visual} shows the visualization results, including input images, facial landmark, segmentation map, attention visualization of ORSANet, and those of the baseline model POSTER~\cite{zheng2023poster}. 
We generate the attention responses map of facial features from the image backbone of the trained model, which maintains the same position as the benchmark model to ensure the fairness of comparison.
The visualization results clearly demonstrate that, under the spatial constraints guided by dense semantics prior and the effective utilization of sparse geometric prior, our ORSANet effectively suppresses interference from non-primary faces (column 1) and occluded regions (column 4) during expression recognition.
In contrast, the baseline model struggles to capture meaningful facial features and exhibits significant activations in occluded areas. 
Furthermore, from the comparison in the fifth column, ORSANet demonstrates stronger responses to key expression-related regions, further validating its superior discriminative ability in real-world complex scenarios.

\textbf{Model Complexity.}
To evaluate the efficiency of our model, 
we provide a comprehensive evaluation of different models in terms of both spatial and temporal complexity, 
as shown in Tab.~\ref{tab:complexity}. Note that we only calculate our trainable part in the whole model. 
The results indicate that our method, with 6.9G FLOPs, is significantly lower than that of most competing methods, being only slightly higher than EAC~\cite{zhang2022learn} while maintaining a moderate parameter count.
This demonstrates that Our ORSANet achieves a balance of performance and computational efficiency.

\begin{table}[t]
\centering
\caption{Comparison of Param and FLOPs with other methods.}
    \label{tab:complexity}
    \renewcommand{\arraystretch}{1.3} 
    \resizebox{1.0\linewidth}{!}{
        \begin{tabular}{l|cc|cc}
        \toprule
        \textbf{Methods} & \textbf{Param (M)} & \textbf{FLOPs (G)} & \textbf{RAF-DB} & \textbf{AffectNet 7cls} \\
        \midrule
        DMUE~\cite{she2021dive}           & 78.4    & 13.4   & 89.42  & 63.11 \\
        TransFER~\cite{xue2021transfer}   & 65.2    & 15.3   & 90.91  & 66.23 \\
        Face2Exp~\cite{zeng2022face2exp}  & \cellcolor{gray!20}\underline{47.1}    & 7.7    & 88.54  & 64.23 \\
        EAC~\cite{zhang2022learn}         & \cellcolor{blue!25}\textbf{25.7}    & \cellcolor{blue!25}\textbf{6.8}    & 90.35  & 65.32 \\
        POSTER~\cite{zheng2023poster}     & 70.1    & 7.4    & 92.05  & 66.17 \\
        CLIPER~\cite{li2024cliper}        & 86.3    & 16.9   & 91.61  & \cellcolor{gray!20}\underline{66.29} \\
        POSTER V2~\cite{mao2024poster++}  & 57.4    & 8.1    & \cellcolor{gray!20}\underline{92.21}  & 66.20  \\
        \midrule
        \textbf{ORSANet} \textit{(Ours)}   & 60.2    & \cellcolor{gray!20}\underline{6.9} & \cellcolor{blue!25}\textbf{92.28}  & \cellcolor{blue!25}\textbf{66.69} \\
        \bottomrule
        \end{tabular}
    }
    \vspace{-0.5em}
\end{table}

\subsection{Ablation study}

\begin{table}[t]
\centering
\caption{Ablation study on the proposed components.}
    \label{tab:ablation}
    \renewcommand{\arraystretch}{1.4} 
    \resizebox{1.0\linewidth}{!}{
        \begin{tabular}{l|ccc}
        \toprule
        \textbf{Variants} & \textbf{RAF-DB} & \textbf{RAF-DB (occlu)} & \textbf{Occlu-FER} \\
        \midrule
        w/o Segmentation map         & 92.15    & 85.44    & 67.27  \\
        w/o Cross-interaction        & 91.99    & 85.35    & 67.32  \\
        w/o Multi-scale              & 92.05    & 85.98    & 67.55  \\
        w/o Landmark Reintegration   & 92.02    & 86.36    & 67.71  \\
        w/o DARELoss                 & \cellcolor{gray!20}\underline{92.22}    & \cellcolor{gray!20}\underline{87.43}    &\cellcolor{gray!20} \underline{67.82}  \\
        \textbf{ORSANet} \textit{(Full)}            & \cellcolor{blue!25}\textbf{92.28}    & \cellcolor{blue!25}\textbf{87.75}    & \cellcolor{blue!25}\textbf{68.07}  \\
        \bottomrule
        \end{tabular}
    }
    \vspace{-1.5em}
\end{table}

\textit{\textbf{Is The Dense Semantics Prior Really Effective?}}
To investigate the effectiveness of the dense semantics prior, we disable the acquisition of facial semantic segmentation information and remove the spatial-semantic guidance module (SSGM). 
As shown in Tab.~\ref{tab:ablation}, although the performance decreases slightly after removing dense semantic prior on RAF-DB, there is a significant decline in the two occlusion-involved test sets, with the model achieving the lowest accuracy of only 67.27\% on Occlu-FER. 
This proves that incorporating dense semantics prior guidance is crucial for semantic reconstruction in occluded regions, enhancing the model's robustness and discriminative capability in real-world occlusion scenarios.

\textit{\textbf{Is The Cross-interaction Mechanism Really Effective?}}
In this experiment, we directly remove the multi-scale cross-interaction module (MCM). 
The results demonstrate that this modification significantly impacts model performance: the recognition accuracy drops substantially across all datasets. 
Without the guidance of landmark information, the network fails to effectively attend to expression-related features, which severely weakens the model's discriminative capability. 
Therefore, the cross-interaction of sparse geometric prior and facial features is essential for ORSANet.

\textit{\textbf{Is The Multi-scale Interaction Really Effective?}}
To validate the effectiveness of multi-scale feature processing, we conduct experiments using only a single-scale CFB. 
Multi-scale interaction not only facilitates the effective processing of expression features at different levels but also meets the fine-grained recognition scale requirements of FER tasks. 
When multi-scale feature extraction is not performed, the model's performance on RAF-DB (occlu) and Occlu-FER decreases by 0.77\% and 0.47\%, respectively. 
This illustrates that single-scale features are insufficient to comprehensively handle multi-level contextual information.

\textit{\textbf{Is The Landmarks Reintegration Mechanism Really Effective?}}
Regarding the reintegration mechanism of landmark information, we remove the adaptive fusion step in each CFB. 
As shown in the experimental results in Tab.~\ref{tab:ablation}, the model accuracy on the natural-scene RAF-DB~\cite{li2017reliable} drops by 0.26\%, which exceeds the removal of dense semantics prior guidance (-0.13\%) and the multi-scale interaction (-0.23\%). 
This indicates that landmark features offer a unique advantage in suppressing intrinsic noises such as identity and illumination variations. 
Therefore, reintegrating landmark information can effectively unleash its sparse potential.

\textit{\textbf{Is DARELoss Really Effective and Can It serves as a Generalized Classification Loss?}}\label{sec:DARELoss}
We conduct ablation studies on the proposed DARELoss, with results presented in Tab.~\ref{tab:ablation}.
On all three validation datasets, the performance of the model decreases to second place after removing DARELoss, showing varying degrees of performance degradation.
This shows the effectiveness of DARELoss in enhancing the learning of complex samples, which can improve the upper bound of the model's performance.
To further verify the generality of DARELoss, we extend our experiments to hyperspectral image classification and natural image classification tasks. To ensure fairness in the experiments, all models used are trained and tested with the publicly available default parameter configurations.
In the hyperspectral image classification task, we integrate DARELoss into the classic framework SpectralFormer~\cite{hong2021spectralformer}. As shown in Tab.~\ref{tab:spectral}, whether on the Indian Pines or Pavia University, adding DARRELoss lead to significant improvements in evaluation metrics for any training strategy.
For natural image classification, we apply DARELoss to two representative architectures: CrossViT~\cite{chen2021crossvit} and GFNet~\cite{rao2021global}, using the CIFAR-10~\cite{krizhevsky2009learning} dataset. As shown in Tab.~\ref{tab:image}, both models benefit from the inclusion of DARELoss, achieving noticeable accuracy gains.
These results clearly demonstrate the versatility and robustness of DARELoss, proving its effectiveness not only in facial expression recognition, but also in broader visual classification tasks.

\begin{table}[t]
\centering
\caption{Performance evaluation of DARELoss integrated into SpectralFormer~\cite{hong2021spectralformer} for hyperspectral image classification. Indian and Pavia Denote Indian Pines dataset and Pavia University dataset, respectively.}
\label{tab:spectral}
    \renewcommand{\arraystretch}{1.3} 
    \resizebox{0.98\linewidth}{!}{
        \begin{tabular}{l|c|c|ccc}
        \hline
        \multirow{2}{*}{\textbf{Datasets}} & \multirow{2}{*}{\textbf{Methods}} & \multirow{2}{*}{\textbf{Config}} & \multicolumn{3}{c}{\textbf{Metrics}} \\ 
        \cline{4-6}
        & & & \textbf{OA (\%)} & \textbf{AA (\%)} & \textbf{\textit{k}} \\ 
        \hline
        \multirow{4}{*}{Indian} 
        & \multirow{2}{*}{pixel-wise} & w/o & \cellcolor{gray!20}74.41 & \cellcolor{gray!20}79.78 & \cellcolor{gray!20}70.24 \\ 
        \cline{3-6}
        & & w/ & \cellcolor{blue!25}\textbf{75.84} & \cellcolor{blue!25}\textbf{82.02} & \cellcolor{blue!25}\textbf{72.08} \\ 
        \cline{2-6}
        & \multirow{2}{*}{patch-wise} & w/o & \cellcolor{gray!20}77.76 & \cellcolor{gray!20}85.13 & \cellcolor{gray!20}74.40 \\ 
        \cline{3-6}
        & & w/ & \cellcolor{blue!25}\textbf{78.63} & \cellcolor{blue!25}\textbf{86.98} & \cellcolor{blue!25}\textbf{75.37} \\ 
        \hline
        
        \multirow{4}{*}{Pavia}
        & \multirow{2}{*}{pixel-wise} & w/o & \cellcolor{gray!20}84.37 & \cellcolor{gray!20}85.35 & \cellcolor{gray!20}78.52 \\ 
        \cline{3-6}
        & & w/ & \cellcolor{blue!25}\textbf{85.78} & \cellcolor{blue!25}\textbf{86.79} & \cellcolor{blue!25}\textbf{81.20} \\ 
        \cline{2-6}
        & \multirow{2}{*}{patch-wise} & w/o & \cellcolor{gray!20}88.40 & \cellcolor{gray!20}88.28 & \cellcolor{gray!20}85.86 \\ 
        \cline{3-6}
        & & w/ & \cellcolor{blue!25}\textbf{88.82} & \cellcolor{blue!25}\textbf{88.74} & \cellcolor{blue!25}\textbf{86.29} \\ 
        \hline
        \end{tabular}
    }
    \vspace{-5pt}
\end{table}

\begin{table}[t]
\centering
\caption{Performance evaluation of DARELoss on CIFAR-10~\cite{krizhevsky2009learning} in a General Image Classification Framework.}
\label{tab:image}
\renewcommand{\arraystretch}{1.3} 
    \resizebox{0.98\linewidth}{!}{
        \begin{tabular}{l|c|cc}
        \hline
        \multirow{2}{*}{\textbf{Methods}} & \multirow{2}{*}{\textbf{Config}} & \multicolumn{2}{c}{\textbf{Metrics}} \\ 
        \cline{3-4}
        & & \textbf{Top-1 ACC (\%)} & \textbf{Top-5 ACC (\%)} \\ 
        \hline 
        \multirow{2}{*}{CrossViT} & w/o & \cellcolor{gray!20}87.91 & \cellcolor{gray!20}98.02 \\ 
        \cline{2-4}
        & w/ & \cellcolor{blue!25}\textbf{88.76} & \cellcolor{blue!25}\textbf{98.43} \\ 
        \hline 
        \multirow{2}{*}{GFNet} & w/o & \cellcolor{gray!20}87.56 & \cellcolor{gray!20}98.14 \\ 
        \cline{2-4}
        & w/ & \cellcolor{blue!25}\textbf{88.13} & \cellcolor{blue!25}\textbf{98.33} \\ 
        \hline
        \end{tabular}
    }
    \vspace{-5pt}
\end{table}

\section{Limitation}
\begin{figure}[t]
    \begin{center}
       \includegraphics[width=1.0\linewidth]{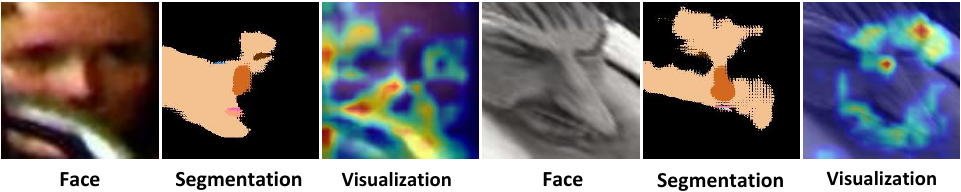} 
    \end{center}
    \caption{Visual examples of deficiencies. When facing low-quality input images or anomalous facial samples, the recognition accuracy of our ORSANet may be limited by the semantic segmentation network.}
    \label{fig:limitation}
    \vspace{-1.0em}
\end{figure}

The image feature processing of ORSANet is limited some extent by the accuracy of dense semantics prior generation, which may result in instability in the extraction of key features. 
Although we employ a pre-trained model as a powerful semantic feature extractor, the current approach may still be insufficient when confronted with low-quality or anomalous facial samples, as shown in Fig.~\ref{fig:limitation}. 
To address these challenges, future research could incorporate more robust semantic segmentation networks and design anomaly detection mechanisms to improve the stability of feature extraction, thereby further enhancing the model's generalization capabilities.

\section{Conclusion}
This paper proposes an Occlusion-Robust Semantic-Aware Network (ORSANet) to address occlusion and dataset imbalance in FER. 
We introduce auxiliary multi-modal semantic guidance that integrates both dense semantics prior (\textit{i.e.}, semantic segmentation maps) and sparse geometric prior (\textit{i.e.}, facial landmarks) to facilitate high-level semantic knowledge learning and disambiguate facial occlusion. A Multi-scale Cross-interaction Module (MCM) is designed to integrate these two priors effectively.
In addition, We propose a Dynamic Adversarial Repulsion Enhancement Loss (DRAELoss) to enhance category discriminability.
Beyond algorithmic innovations, we further construct the Occlu-FER dataset, specialized for occluded scenes, to evaluate the model's robustness under various real-world occlusions.
Extensive experiments demonstrate that our ORSANet achieves SOTA performance.

\section*{Acknowledgements}
This work is supported in part by the National Natural Science Foundation of China (No. U2333211), in part by the Fundamental Research Funds for the Central Universities (No. ZYGX2024Z004), and in part by the Project of Sichuan Engineering Technology Research Center for Civil Aviation Flight Technology and Flight Safety (No. GY2024-27D).

\bibliographystyle{ACM-Reference-Format}
\balance
\bibliography{acmart}

\end{document}